\documentclass{article}

\usepackage[main, final]{neurips_2026}
\setcitestyle{numbers,square,comma,sort}
\usepackage{graphicx}
\usepackage{multirow}
\usepackage{amsmath}
\usepackage{amssymb}
\usepackage{pifont}
\usepackage{soul}
\usepackage{siunitx}
\newcommand{\cmark}{\ding{51}}  
\newcommand{\xmark}{\ding{55}}  

\usepackage[utf8]{inputenc} 
\usepackage[T1]{fontenc}    
\usepackage{hyperref}       
\usepackage{url}            
\usepackage{booktabs}       
\usepackage{amsfonts}       
\usepackage{nicefrac}       
\usepackage{microtype}      
\usepackage{xcolor}         
\usepackage[table]{xcolor}
\definecolor{yellow}{rgb}{1, 1, 0.7}
\definecolor{orange}{rgb}{1, 0.85, 0.7}
\definecolor{red1}{rgb}{1, 0.7, 0.7}
\title{T$^2$-Splat: Adaptive Topology Mesh Splatting with Texture Residuals}

\author{
  Zhihao Tang$^{1}$ \quad
  Youjia Zhang$^{1}$ \quad
  Mingbo Zhao$^{2}$ \quad
  Wei Yang$^{1}$\thanks{
    Corresponding author: \texttt{weiyangcs@hust.edu.cn}.
  } \\
  $^{1}$Huazhong University of Science and Technology \\
  $^{2}$Donghua University
}

\begin{document}

\setcounter{footnote}{1}
\maketitle
\setcounter{footnote}{0}

\vspace{-20pt}
\begingroup
\setlength{\intextsep}{4pt}

\begin{figure}[h!]
  \centering
  \includegraphics[width=1.0\textwidth]{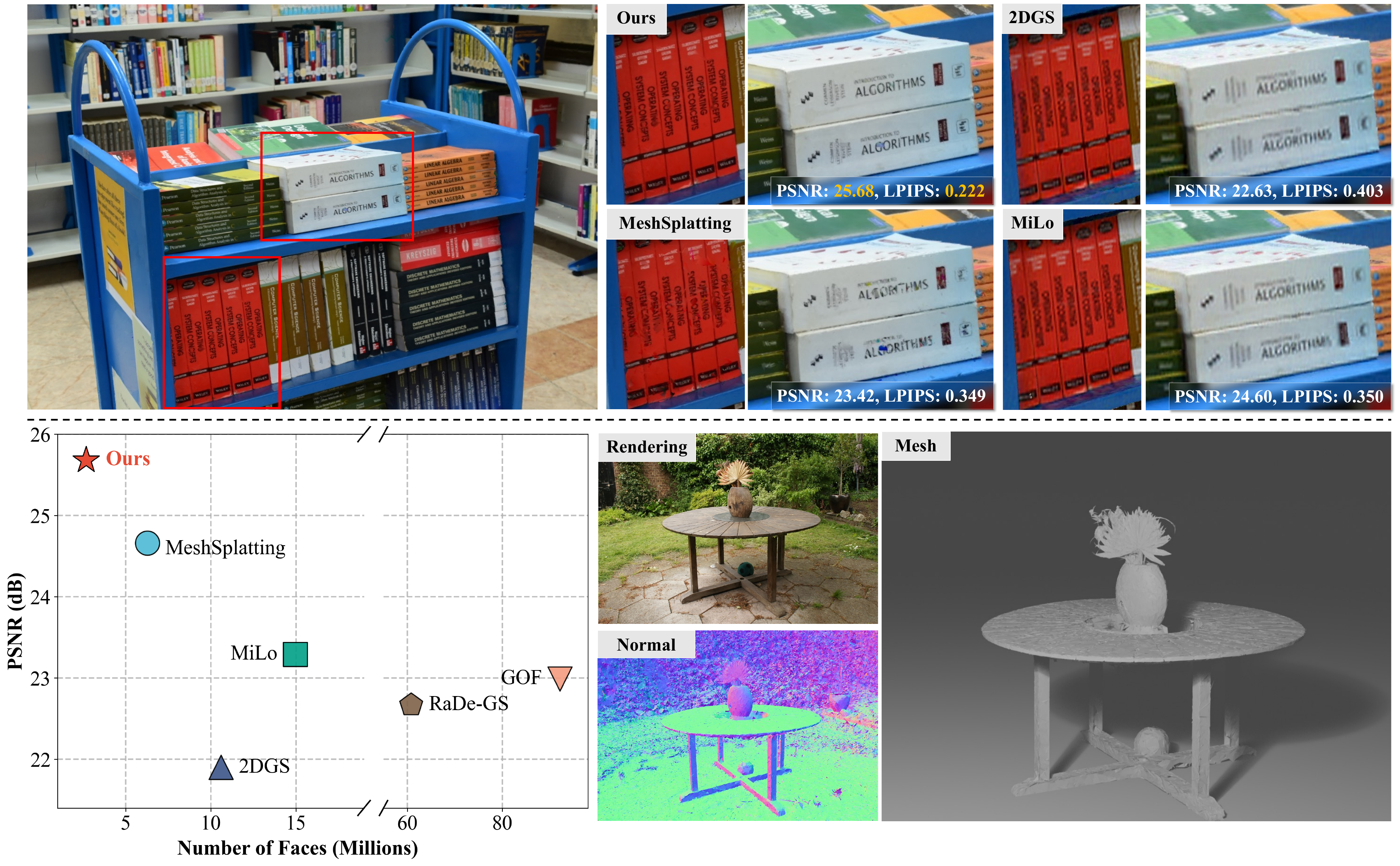}
  \caption{
Top: our method better preserves high-frequency details
(e.g., text) than MeshSplatting~\cite{Held2025MeshSplatting}
and MiLo~\cite{guedon2025milo} on the \textit{shelf} scene
(STRinGS-360~\cite{STRinGS_2026_WACV}).
Bottom: on the \textit{garden} scene, T$^2$-Splat achieves
the highest PSNR (25.68\,dB) among the compared methods
using only 2.7M faces, versus 92M for GOF~\cite{Yu2024GOF}
and 15M for MiLo~\cite{guedon2025milo}.
}
\end{figure}

\endgroup

\begin{abstract}
Triangle-based splatting connects novel view synthesis with
conventional graphics pipelines, but often requires excessive
tessellation to preserve appearance details.
We present T$^2$-Splat, which decouples topology adaptation from
appearance refinement. Safe Restricted Delaunay Triangulation (Safe RDT) initializes
mesh connectivity. Dynamic Topology Evolution (DTE) then adapts
triangle allocation through edge collapse and vertex split operations.
Residual per-face Mip-textures then refine appearance on the
fixed mesh using scale-adaptive filtering.
Experiments on Mip-NeRF 360, Tanks \& Temples, and DTU demonstrate
competitive results with compact meshes.
T$^2$-Splat improves PSNR by $0.71$\,dB over MeshSplatting on
Mip-NeRF 360. Additional compression experiments demonstrate flexible control of mesh complexity, supporting face budgets as low as 10\% of the original face count.
Project page: \url{https://t2-splat.github.io/}.

\end{abstract}

\section{Introduction}

3D Gaussian Splatting (3DGS) has enabled high-fidelity and real-time novel view synthesis by representing scenes as sets of anisotropic Gaussian primitives~\cite{kerbl3Dgaussians}. 
Despite its impressive rendering quality and optimization efficiency, its unstructured point-based representation remains difficult to integrate into standard graphics pipelines. 
In particular, 3DGS relies on visibility sorting and alpha compositing, which are not naturally aligned with conventional rasterization mechanisms such as depth buffering, hardware occlusion culling, and mesh-based asset processing~\cite{akenine2019real,foley1996computer}. 
This mismatch complicates deployment in game engines, AR/VR systems, and interactive graphics applications, where triangle meshes remain the dominant representation.

Motivated by this gap, recent work has been resorting to triangle- and mesh-based representations for differentiable novel view synthesis. 
A line of methods converts Gaussian or neural representations into meshes through post-processing~\cite{Huang2DGS2024,Yu2024GOF,zhang2026rade,guedon2024sugar}, but such two-stage pipelines often decouple rendering optimization from the final mesh structure, leading to detail loss or additional engineering complexity. 
More recent opaque triangle-based splatting methods directly optimize triangle primitives within differentiable rendering frameworks, improving compatibility with real-time graphics systems while retaining competitive visual quality. 
For example, MeshSplatting~\cite{Held2025MeshSplatting} jointly optimizes geometry and appearance with connected opaque meshes and restricted Delaunay triangulation, demonstrating the promise of mesh-native splatting for efficient novel view synthesis. 
However, existing mesh splatting frameworks still face an important limitation: the allocation of triangle primitives is not sufficiently adaptive to the spatially varying complexity of real scenes.

This limitation arises because triangle primitives carry a dual responsibility. 
On the one hand, they define the geometric support of the reconstructed scene; on the other hand, they also provide the basis on which appearance features are stored and interpolated. 
When the mesh contains too many triangles in smooth or textureless regions, the representation becomes unnecessarily redundant and inefficient. 
Conversely, when the triangle budget is reduced, high-frequency appearance details may be lost, as vertex-based appearance models often require dense tessellation to reproduce fine texture variations. 
As a result, topology simplification and appearance fidelity become tightly coupled: aggressively collapsing triangles can improve compactness, but may blur textures, introduce aliasing, or degrade view-dependent details. 
A more desirable mesh splatting representation should therefore adapt its topology according to geometric complexity while providing an appearance model that remains expressive under changing triangle resolutions.

To address this problem, we propose \textbf{T$^2$-Splat}, a mesh splatting framework with adaptive topology and texture optimization. 
The key idea is to make triangle allocation and appearance modeling mutually supportive rather than restrictive. 
At the geometry level, we introduce a dynamic topology optimization strategy based on progressive edge-collapse and vertex-split operations~\cite{hoppe1996progressive}. Edge collapse removes redundant triangles in locally simple regions, while vertex split restores degrees of freedom in regions requiring finer geometric or appearance detail. These local and reversible operations allow the mesh to adapt its connectivity during optimization, reducing unnecessary face counts while preserving well-conditioned manifold structure. At the appearance level, we introduce a scale-aware residual mip-texture model. 
Instead of relying solely on vertex-interpolated appearance features, we attach residual texture signals to triangle primitives and organize them in a mip-texture hierarchy~\cite{ewins1998mip}. 
DTE first adapts mesh connectivity, after which residual
per-face textures and vertex SH coefficients are jointly
optimized on the fixed mesh to recover appearance details.
This separates topology adaptation from subsequent
appearance refinement.
Moreover, LOD-aware texture sampling provides scale-dependent filtering, reducing aliasing and preserving visual stability across camera distances and viewpoints~\cite{luebke2002level}. 
Together, adaptive topology evolution and residual mip-texture modeling enable T$^2$-Splat to achieve a better balance between compact mesh structure and high-fidelity rendering.
We evaluate T$^2$-Splat on the Mip-NeRF 360, Tanks \& Temples, and DTU benchmarks. 
Experiments show that our method improves novel-view rendering quality by +0.71 dB over MeshSplatting on Mip-NeRF 360, while maintaining robust visual fidelity even when the mesh is compressed to 10\% of the original face count. 

In summary, our main contributions are:

\begin{itemize}
    \item \textbf{T$^2$-Splat}, a mesh splatting framework featuring a decoupled topology and texture optimization strategy, enabling compact yet high-fidelity novel view synthesis.

    \item We demonstrate that progressive mesh operations, i.e., edge collapse and vertex split, can be naturally integrated into a differentiable rendering pipeline, for reallocating triangle primitives according to scene complexity while preserving well-conditioned mesh connectivity.

    \item A scale-aware residual Mip-texture model for triangle primitives that decouples high-frequency appearance from mesh density, improving rendering fidelity under topology simplification.
\end{itemize}

\section{Related Work}
\label{gen_inst}

\textbf{Mesh Extraction from Radiance Field.} Neural Radiance Fields (NeRF)\cite{mildenhall2021nerf} pioneered photorealistic novel view synthesis, but dense ray sampling incurs high rendering costs. Subsequently, 3D Gaussian Splatting (3DGS)\cite{kerbl3Dgaussians} achieved real-time rendering using discrete point-based primitives; however, its unstructured, alpha-blended nature is inherently incompatible with standard graphics pipelines. To bridge this gap, early methods\cite{yariv2023bakedsdf,chen2023mobilenerf,reiser2024binary,yariv2021volume,wang2021neus} extracted meshes from implicit fields, but this introduced overhead and increased the overall training time. For 3DGS, recent works such as 2DGS\cite{Huang2DGS2024} and RaDe-GS\cite{zhang2026rade} utilize Truncated Signed Distance Fields (TSDF)\cite{curless1996volumetric} for surface extraction, while SuGaR\cite{guedon2024sugar} applies Poisson Reconstruction\cite{kazhdan2006poisson} by sampling a surface-aligned Gaussian level set. Additionally, MiLo\cite{guedon2025milo} introduces a hybrid framework utilizing Differentiable Marching Tetrahedra (DMTet)\cite{shen2021deep} to dynamically recover meshes. However, because these methods still inherently rely on implicit fields or independent post-processing steps , this decoupling between the extraction phase and the optimization phase often leads to a noticeable loss of fine geometric details. 

\textbf{Differentiable Mesh Reconstruction.} To avoid precision loss from format conversion, direct end-to-end optimization of meshes has emerged. Early attempts\cite{son2024dmesh,son2025dmesh++} at differentiable mesh reconstruction were primarily limited to simple synthetic objects and often struggled to generalize to real-world complex scenes. Existing methods\cite{jiang2025halogs,Held2025Triangle,burgdorfer2025radiant} typically optimize an unstructured collection of triangles, resulting in a lack of connectivity between adjacent triangles. Furthermore, they often fail to generate opaque triangles at the end of training, which limits their usability in downstream applications. While methods like Triangle Splatting+\cite{held2025triangle+} and MeshSplatting\cite{Held2025MeshSplatting} achieve connected meshes via Restricted Delaunay Triangulation (RDT)\cite{cheng2013delaunay} and successfully generate opaque triangles, they suffer from severe Topological Rigidity: once established, the mesh topology is frozen, preventing adaptation to scene complexity. In contrast, T$^2$-Splat integrates Progressive Meshes into a differentiable pipeline, utilizing explicit edge collapse and vertex split for dynamic geometric complexity adaptation.  

\textbf{Appearance Modeling.} Achieving high-fidelity appearance on meshes is challenging. Most mesh-based baselines\cite{Held2025Triangle,Held2025MeshSplatting} and early 3DGS methods rely on Spherical Harmonics (SH)\cite{green2003spherical} attached to vertices. This assumes smooth transitions and severely limits the recovery of high-frequency details. Inspired by recent texture-based Gaussian splatting methods\cite{rong2024gstex,svitov2025billboard,wang2025neural,xu2024texture,chao2025texturedgaussians}, we discard the vertex-only appearance model. Instead, we introduce a Residual Per-face Texture to recover high-frequency spatial details. Furthermore, we integrate Level-of-Detail (LOD)\cite{luebke2002level} and Mip-texture pyramids\cite{ewins1998mip}, enabling scale-adaptive trilinear sampling to effectively eliminate texture aliasing and flickering across viewing distances.

\section{Method}
\label{headings}
The overall architecture of T$^2$-Splat consists of two main stages. 
In the geometry stage, we construct an initial manifold mesh using Safe Restricted Delaunay Triangulation (Safe RDT)~\cite{cheng2013delaunay} following MeshSplatting~\cite{Held2025MeshSplatting} and progressively refine its structure through periodic topology editing. 
Specifically, we apply edge-collapse and vertex-split operations to adaptively reallocate triangle primitives according to local scene complexity while preserving valid mesh connectivity. 
In the appearance stage, we introduce zero-initialized residual per-face textures organized in a mip-texture pyramid~\cite{ewins1998mip}. 
This design injects high-frequency appearance details without perturbing the optimized geometry, enabling scale-aware texture filtering and anti-aliased rendering.

\begin{figure}[t!]
  \centering
  \includegraphics[width=1.0\textwidth]{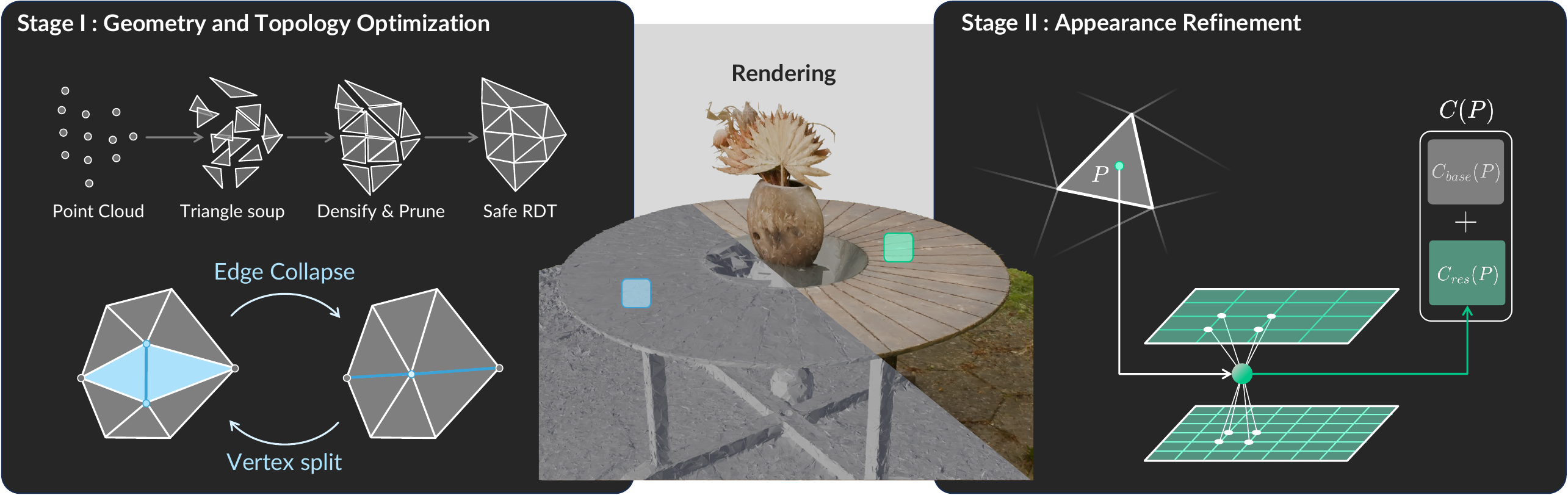}
  \caption{$T^2$-Splat is an end-to-end differentiable rendering framework that bridges high-fidelity novel view synthesis and standard graphics pipelines. In Stage I, it constructs a manifold mesh from an unstructured triangle soup via Safe Restricted Delaunay Triangulation (Safe RDT), and introduces Dynamic Topology Evolution (DTE) to adaptively optimize topology using QEM-based edge collapses and gradient-driven vertex splits. Stage II employs a scale-aware residual Mip-texture model with trilinear sampling, decoupling appearance from geometry to ensure high-quality, anti-aliased rendering even under extreme mesh compression.}
  \label{fig:pipelime}
\end{figure}
\subsection{Preliminary}

We follow the definition of opaque triangle primitives in MeshSplatting~\cite{Held2025MeshSplatting}, i.e., where each vertex of a triangle stores position $v_i$, opacity logit $a_i$, and spherical harmonics coefficients $c_i$. For a pixel $p$ inside a triangle, its color is obtained by barycentric interpolation of vertex attributes:
\begin{equation}
C_{\text{base}}(p, d) = w_i c_i(d) + w_j c_j(d) + w_k c_k(d), \quad w_i+w_j+w_k=1
\end{equation}
where $d$ denotes the viewing direction. While this enables real-time rendering, the spatial resolution of the appearance is strictly tied to the vertex density.

To make triangle rasterization end-to-end differentiable, the discrete triangle indicator is replaced by a soft window function $I(p)$:
\begin{equation}
I(p) = \left(\operatorname{ReLU}\left(\frac{\phi(p)}{\phi(s)}\right)\right)^\sigma
\end{equation}
Here, $\phi(p)$ represents the signed distance field of the 2D triangle in image space, which takes negative values inside, positive outside, and zero on the boundary. The vector $s$ is the incenter of the projected triangle, ensuring that the fraction evaluates to $1$ at the center and $0$ at the edges. 

The exponent $\sigma$ is a smoothness parameter that controls the transition profile. Concurrently, the rendering opacity $\alpha_i$ is re-parameterized as:
\begin{equation}
\alpha_i = \alpha_{\text{floor}} + (1-\alpha_{\text{floor}}) \cdot \operatorname{sigm}(a_i)
\label{eq_a}
\end{equation}
Visible triangles are then alpha-blended in depth order to compute the final baseline pixel color $C_{\text{base}}(p, d)$:
\begin{equation}
C_{\text{base}}(p, d) = \sum_{n=1}^{N} c_{T_n}(p, d)\alpha_{T_n}I(p)\prod_{m=1}^{n-1} \left(1-\alpha_{T_m}I(p)\right)
\end{equation}
During training, $\sigma$ is linearly annealed from $1.0$ (a smooth gradient facilitating gradient flow) down to $0.0001$. As $\alpha_{\text{floor}}$ approaches $1$ and $\sigma$ approaches $0$, the representation smoothly converges to solid, opaque triangles. Analogously to MobileNeRF~\cite{chen2023mobilenerf}, this formulation simplifies at the end of training to:$C_{\text{base}}(p, d) = c_{T_n}(p, d) I(p)$,where only a single evaluation per pixel is required. This reduces over-drawing to zero, significantly accelerating the rendering process and enabling native compatibility with standard game engines. This process generates redundant triangles, while geometrically complex regions may lack sufficient degrees of freedom. In the following, we propose to use the progressive mesh operations, i.e., vertex splitting and edge collapsing, to optimize the topology.

\begin{figure}[t!]
  \centering
  \includegraphics[width=1.0\textwidth]{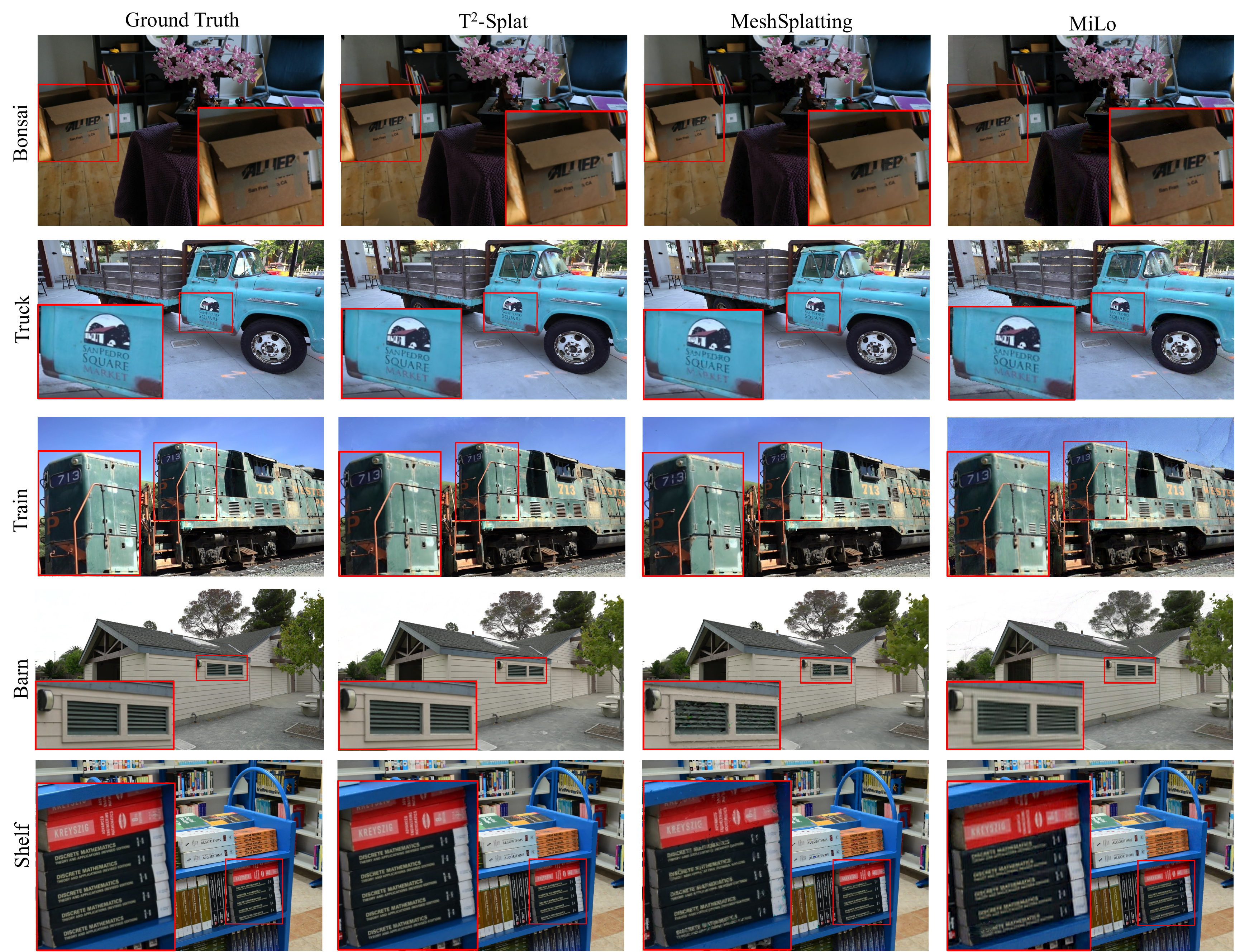}
  \caption{Qualitative results. Compared to baselines that suffer from blurry textures or geometric distortions, our T$^2$-Splat utilizes per-face textures to effectively eliminate texture distortion and preserves significantly sharper high-frequency details (e.g., fine text and intricate surface patterns) and maintains better structural integrity.}
  \label{fig:visual}
\end{figure}
\subsection{Geometry and Topology Optimization}

\subsubsection{Safe Restricted Delaunay Triangulation.}
To initialize DTE, we augment Restricted Delaunay Triangulation
(RDT) with a constrained face-selection procedure, termed
\emph{Safe RDT}. Starting from the candidates retained by the
RDT intersection test, we filter and rank faces according to
their distance to the input triangle soup, unsigned normal
agreement, and triangle quality. We then greedily grow a
consistently oriented surface along its edge frontier, ensuring
that no duplicate faces are introduced, each edge has at most
two incident faces, and each used vertex has a single connected
face fan. This procedure updates only face connectivity and
winding, retaining vertex positions, appearance and opacity
parameters, and their optimizer states. Open boundaries and
multiple connected components are permitted. The selected mesh
is subsequently refined by DTE through edge collapse and
vertex split operations. Further details are provided in the supplementary material.

\subsubsection{Edge Collapse: QEM-based Geometric Simplification}

To remove redundant triangles in flat regions while preserving boundaries, we use Quadric Error Metrics (QEM). For each face with plane equation $\mathbf{p}_f=[a_f,b_f,c_f,d_f]^T$, its quadric is $K_f=\mathbf{p}_f\mathbf{p}_f^T$. Vertex quadrics are accumulated from adjacent faces (with additional virtual planes for open boundaries):
\begin{equation}
Q_v=\sum_{f\in\mathcal{F}(v)}K_f.
\end{equation}
For a candidate edge $e=(u,v)$, the merged quadric is $Q_{uv}=Q_u+Q_v$. The optimal new vertex position minimizes the homogeneous quadric error:
\begin{equation}
\bar{v}^{*} = \arg\min_{\bar{v}} \bar{v}_h^T Q_{uv}\bar{v}_h, \quad \bar{v}_h=[x,y,z,1]^T.
\end{equation}
If the system is singular, we fall back to the best discrete candidate:
\begin{equation}
\bar{v}^{*} = \arg\min_{c\in\{u,v,(u+v)/2\}} c_h^T Q_{uv} c_h.
\end{equation}
The collapse cost combines the normalized QEM error with
base-color and activated opacity differences to discourage
collapses across appearance and opacity discontinuities:
\begin{equation}
    \operatorname{Cost}(e)
    =
    w_q
    \frac{
        (\bar{\mathbf{v}}_h^{*})^{\top}
        Q_{uv}
        \bar{\mathbf{v}}_h^{*}
    }{
        \operatorname{median}(E_q) + \epsilon
    }
    +
    w_c
    \left\|
        \mathbf{c}_u^{\mathrm{dc}}
        -
        \mathbf{c}_v^{\mathrm{dc}}
    \right\|_2
    +
    w_a \left| \alpha_u - \alpha_v \right|.
    \label{eq:collapse_cost}
\end{equation}
Here, $\alpha_i$ denotes the activated vertex opacity
defined in Eq.~\eqref{eq_a} and $w_a$ controls the opacity term.
After sorting candidate edges by cost, we select collapses
subject to a maximum edge-length threshold and the full
simplicial link condition:
\begin{equation}
\operatorname{Lk}_{\mathcal{M}}(u)
\cap
\operatorname{Lk}_{\mathcal{M}}(v)
=
\operatorname{Lk}_{\mathcal{M}}(\{u,v\}).
\label{eq:link_condition}
\end{equation}
Here, $\operatorname{Lk}_{\mathcal{M}}(\cdot)$ denotes the
simplicial link in the current mesh $\mathcal{M}$, including
its vertices and edges.
We additionally reject collapses that join two boundary
vertices through an interior edge, produce an invalid merged
vertex fan, or introduce local face degeneracy or orientation
reversal. Accepted collapses within each update have disjoint
affected one-ring neighborhoods.

Appearance attributes for the new vertex are merged via opacity-weighted averaging:
\begin{equation}
c_{new}= \frac{\alpha_u c_u+\alpha_v c_v}{\alpha_u+\alpha_v+\epsilon}.
\end{equation}

\subsubsection{Vertex Splitting: Gradient-driven Refinement}

To increase local mesh density in complex regions, we calculate a perspective-compensated gradient score using the position gradient magnitude and camera-space depth $d_i$:
\begin{equation}
g_i^c=\|\nabla_{v_i}\mathcal{L}\|\cdot d_i.
\end{equation}
This score is accumulated and normalized across $N_i$ training views:
\begin{equation}
\bar{g}_i^c= \frac{1}{N_i} \sum_{k=1}^{N_i}g_{i,k}^c.
\end{equation}

Vertices with high $\bar{g}_i^c$ are refined by inserting a new vertex at the midpoint of their longest adjacent edge:
\begin{equation}
v_{new}=\frac{1}{2}(v_i+v_j).
\end{equation}
The incident triangle is replaced by two new triangles while preserving orientation:
\begin{equation}
(i,j,k)\rightarrow(i,v_{new},k),(v_{new},j,k).
\end{equation}
Finally, the attributes of the inserted vertex are linearly interpolated:
\begin{equation}
c_{new}=\frac{c_i+c_j}{2}, \quad a_{new}=\frac{a_i+a_j}{2}.
\end{equation}

\subsubsection{Optimizer State Stitching}

To preserve optimization history across edge collapse and
vertex split operations, we transfer Adam's first and second
moments, denoted by $m$ and $v$, to the updated parameters.

During edge collapse, the merged vertex inherits the moments
of the two endpoints through incident-area-weighted averaging:
\begin{equation}
m_{\mathrm{new}}
=
\frac{A_u m_u+A_v m_v}{A_u+A_v},
\qquad
v_{\mathrm{new}}
=
\frac{A_u v_u+A_v v_v}{A_u+A_v},
\label{eq:collapse_optimizer_state}
\end{equation}
where $A_u$ and $A_v$ are the incident-area weights of
vertices $u$ and $v$.

During vertex splitting, the retained parent $k$ and the newly
inserted midpoint vertex $k'$ inherit the parent's moments,
with opposite perturbations applied to the first moment:
\begin{equation}
\begin{aligned}
m_k^{+} &= m_k+\Delta m,
\qquad m_{k'}=m_k-\Delta m,\\
v_k^{+} &= v_{k'}=v_k.
\end{aligned}
\label{eq:split_optimizer_state}
\end{equation}
Here, $+$ denotes the updated state of the retained vertex,
$\Delta m=10^{-3}\sqrt{v_k}\odot z$,
and $z\sim\mathcal{N}(0,I)$.
The square root and multiplication are element-wise.

\begin{table}[t]
    \centering

    \caption{\textbf{\ul{Mesh-based} NVS results on Mip-NeRF360 and Tanks \& Temples.}
    We report PSNR, SSIM, LPIPS, and vertex count.
    \emph{Mesh}, \emph{Color}, and \emph{Connect} indicate whether the method
    directly outputs a mesh, whether the mesh carries color without
    post-processing, and whether the output is connected, respectively.}
    \label{tab:comparisons_main_table}
    \par\vspace{4pt}

    \footnotesize\rmfamily
    \setlength{\tabcolsep}{2pt}
    \renewcommand{\arraystretch}{1}

    \begin{tabular*}{\linewidth}{@{\extracolsep{\fill}}l|ccc|cccc|cccc@{}}
        \multirow{2}{*}{Method}
        & \multicolumn{3}{c}{Characteristics}
        & \multicolumn{4}{c}{Mip-NeRF360 dataset}
        & \multicolumn{4}{c}{Tanks \& Temples} \\
        & Mesh & Color & Connect
        & PSNR $\uparrow$
        & LPIPS $\downarrow$
        & SSIM $\uparrow$
        & \#Verts $\downarrow$
        & PSNR $\uparrow$
        & LPIPS $\downarrow$
        & SSIM $\uparrow$
        & \#Verts $\downarrow$ \\
        \midrule

        3DGS & -- & -- & --
            & 27.21 & 0.214 & 0.815 & --
            & 23.14 & 0.183 & 0.841 & -- \\

        Triangle Splatting & -- & -- & --
            & 27.16 & 0.191 & 0.814 & --
            & 23.14 & 0.143 & 0.857 & -- \\

        \midrule
        2DGS & \xmark & \xmark & \cmark
            & 15.36 & 0.474 & 0.498 & \cellcolor{red1}2M
            & 14.23 & 0.485 & 0.569 & 16M \\

        GOF & \xmark & \xmark & \cmark
            & 20.78 & 0.465 & 0.573 & 33M
            & \cellcolor{red1}21.69
            & \cellcolor{yellow}0.326
            & 0.690 & 12M \\

        RaDe-GS & \xmark & \xmark & \cmark
            & 23.56 & 0.361 & 0.668 & 31M
            & 20.51 & 0.344 & 0.659 & 10M \\

        MiLo & \cmark & \xmark & \cmark
            & 24.09 & 0.323 & 0.688 & \cellcolor{yellow}7M
            & \cellcolor{orange}21.46
            & 0.348
            & \cellcolor{yellow}0.706
            & \cellcolor{yellow}4M \\

        Gaussian Wrapping & \cmark & \xmark & \cmark
            & \cellcolor{orange}25.10
            & \cellcolor{orange}0.289
            & \cellcolor{orange}0.752
            & 12M
            & 20.98 & 0.367 & 0.700 & 6M \\

        Triangle Splatting & \cmark & \cmark & \xmark
            & 21.05 & 0.462 & 0.558 & \cellcolor{orange}3M
            & 17.27 & 0.402 & 0.600 & 6M \\

        MeshSplatting & \cmark & \cmark & \cmark
            & \cellcolor{yellow}24.78
            & \cellcolor{yellow}0.310
            & \cellcolor{yellow}0.728
            & \cellcolor{orange}3M
            & 20.52
            & \cellcolor{orange}0.287
            & \cellcolor{orange}0.745
            & \cellcolor{orange}2M \\

        \midrule
        \textbf{T$^2$-Splat} & \cmark & \cmark & \cmark
            & \cellcolor{red1}25.49
            & \cellcolor{red1}0.234
            & \cellcolor{red1}0.779
            & \cellcolor{red1}2M
            & \cellcolor{yellow}21.12
            & \cellcolor{red1}0.213
            & \cellcolor{red1}0.795
            & \cellcolor{red1}1M \\
    \end{tabular*}
\end{table}

\begin{table}[t]
    \centering

    \caption{\textbf{Chamfer distance on the DTU dataset.}
    We report Chamfer distance across 15 scenes.
    T$^2$-Splat achieves a joint second-best mean Chamfer distance of 0.68
    and the lowest error on three scenes (63, 83, and 105).}
    \label{tab:dtu_sum}
    \par\vspace{4pt}

    \footnotesize\rmfamily
    \setlength{\tabcolsep}{2pt}
    \renewcommand{\arraystretch}{1}

    \begin{tabular*}{\linewidth}{@{\extracolsep{\fill}}l|*{15}{c}|c@{}}
        Method
            & 24 & 37 & 40 & 55 & 63
            & 65 & 69 & 83 & 97 & 105
            & 106 & 110 & 114 & 118 & 122 & Mean \\
        \midrule

        3DGS
            & 2.14 & 1.53 & 2.08 & 1.68 & 3.49
            & 2.21 & 1.43 & 2.07 & 2.22 & 1.75
            & 1.79 & 2.55 & 1.53 & 1.52 & 1.50 & 1.96 \\

        2DGS
            & 0.48 & 0.91 & 0.39 & 0.39 & 1.01
            & 0.83 & 0.81 & 1.36 & 1.27 & 0.76
            & 0.70 & 1.40 & 0.40 & 0.76 & 0.52 & 0.80 \\

        GOF
            & 0.50 & 0.82 & 0.37
            & \cellcolor{orange}0.37
            & 1.12
            & \cellcolor{orange}0.74
            & 0.73 & 1.18 & 1.29 & 0.68
            & 0.77 & 0.90 & 0.42
            & \cellcolor{yellow}0.66
            & 0.49
            & \cellcolor{yellow}0.74 \\

        RaDe-GS
            & \cellcolor{yellow}0.46
            & 0.73
            & \cellcolor{orange}0.33
            & \cellcolor{yellow}0.38
            & \cellcolor{orange}0.79
            & \cellcolor{yellow}0.75
            & 0.76 & 1.19 & 1.22 & 0.62 & 0.70
            & \cellcolor{yellow}0.78
            & \cellcolor{orange}0.36
            & 0.68
            & \cellcolor{orange}0.47
            & \cellcolor{orange}0.68 \\

        MiLo
            & \cellcolor{orange}0.43
            & 0.74
            & \cellcolor{yellow}0.34
            & \cellcolor{orange}0.37
            & \cellcolor{yellow}0.80
            & \cellcolor{orange}0.74
            & \cellcolor{yellow}0.70
            & 1.21 & 1.22 & 0.66
            & \cellcolor{orange}0.62
            & 0.80
            & \cellcolor{yellow}0.37
            & 0.76
            & \cellcolor{yellow}0.48
            & \cellcolor{orange}0.68 \\

        Triangle Splatting
            & 0.98 & 1.07 & 1.07 & 0.51 & 1.67
            & 1.44 & 1.17 & 1.32 & 1.75 & 0.98
            & 0.96 & 1.11 & 0.56 & 0.93 & 0.72 & 1.08 \\

        MeshSplatting
            & 0.77
            & \cellcolor{yellow}0.72
            & 0.87 & 0.86 & 0.81 & 1.35 & 0.89
            & \cellcolor{yellow}1.08
            & \cellcolor{orange}1.15
            & \cellcolor{yellow}0.58
            & 0.95 & 1.00 & 0.75 & 0.86 & 0.78 & 0.89 \\

        Gaussian Wrapping
            & \cellcolor{red1}0.40
            & \cellcolor{red1}0.51
            & \cellcolor{red1}0.27
            & \cellcolor{red1}0.32
            & 0.81
            & \cellcolor{red1}0.46
            & \cellcolor{red1}0.45
            & \cellcolor{orange}1.04
            & \cellcolor{red1}0.65
            & \cellcolor{orange}0.52
            & \cellcolor{red1}0.33
            & \cellcolor{red1}0.59
            & \cellcolor{red1}0.32
            & \cellcolor{red1}0.33
            & \cellcolor{red1}0.34
            & \cellcolor{red1}0.49 \\

        \midrule
        \textbf{T$^2$-Splat}
            & 0.56
            & \cellcolor{orange}0.63
            & 0.55 & 0.47
            & \cellcolor{red1}0.77
            & 1.09
            & \cellcolor{orange}0.69
            & \cellcolor{red1}0.98
            & \cellcolor{yellow}1.18
            & \cellcolor{red1}0.47
            & \cellcolor{yellow}0.63
            & \cellcolor{orange}0.75
            & 0.47
            & \cellcolor{orange}0.53
            & \cellcolor{yellow}0.48
            & \cellcolor{orange}0.68 \\
    \end{tabular*}
\end{table}

\subsection{Textural Residual Model}

With well-conditioned topology, we freeze the mesh topology and vertex positions to focus on appearance refinement. While vertex-based Spherical Harmonics (SH) capture smooth view-dependent effects, they struggle with high-frequency details. We therefore introduce a learnable residual texture $T_f \in \mathbb{R}^{R \times R \times 3}$ for each face $f$, initialized to zero to ensure a lossless transition from Stage I.

\subsubsection{Residual Parameterization and Mirror Mapping}
For a pixel $p$ within face $f$, we define local texture coordinates $(u, v)$ via its barycentric coordinates $(w_i, w_j, w_k)$ such that $u=w_j, v=w_k$. The final color $C(p, d)$ integrates vertex-based shading and per-face residual sampling:

\begin{equation}
C(p, d) = \underbrace{C_{base}(p, d)}_{\text{View-dependent Base}} + \underbrace{C_{res}(u,v)}_{\text{Spatial Residual Texture}},
\end{equation}
where $d$ denotes the viewing direction and $C_{res}(u,v) = \operatorname{Bilinear}(T_f, u, v)$. 
To efficiently utilize the square texture domain and stabilize boundary interpolation, we employ \textit{mirror mapping}: if $u+v>1$, the coordinates are reflected back into the valid triangular domain via $T_f(u,v)=T_f(1-v,1-u)$. This symmetry prevents sampling artifacts and ensures gradient consistency during optimization.

\subsubsection{Scale-adaptive Trilinear Sampling}
To mitigate aliasing caused by small texture footprints, we construct a differentiable Mip-texture pyramid where coarser levels are generated by average pooling: $T_f^{l+1}=\operatorname{AvgPool}(T_f^l)$. The level of detail $\lambda$ is estimated from screen-space derivatives:
\begin{equation}
\lambda= \log_2 \left( \max \left( \sqrt{ \left(\frac{\partial u}{\partial x}\right)^2+ \left(\frac{\partial v}{\partial x}\right)^2 }, \sqrt{ \left(\frac{\partial u}{\partial y}\right)^2+ \left(\frac{\partial v}{\partial y}\right)^2 } \right) +\epsilon \right).
\end{equation}
Defining $D=\lfloor \lambda \rfloor$ and $\delta=\lambda-D$, the final texture value $C_{res}^\lambda(u, v)$ is computed via trilinear interpolation between adjacent Mip levels:

\begin{equation}
C_{res}^\lambda(u, v) = (1-\delta)\operatorname{Bilinear}(T_f^D,u,v) + \delta\operatorname{Bilinear}(T_f^{D+1},u,v).
\end{equation}
The total color is thus refined as $C(p, d) = C_{base}(p, d) + C_{res}^\lambda(u, v)$, effectively acting as a dynamic low-pass filter to eliminate flickering while preserving high-frequency details across varying scales.

\subsection{Optimization Objectives}

In the geometric stage, we optimize vertex positions $v_i$, opacities $o_i$, and spherical harmonic (SH) coefficients $c_i$ with a joint objective:
\begin{equation}
\mathcal{L} = \mathcal{L}_{\text{3DGS}} + \beta_{o}\mathcal{L}_{o} + \beta_{z}\mathcal{L}_{z} + \beta_{n}\mathcal{L}_{n} + \beta_{d}\mathcal{L}_{d}.
\end{equation}
Here, $\mathcal{L}_{\text{3DGS}}$ combines the photometric $\mathcal{L}_1$ and $\mathcal{L}_{\text{D-SSIM}}$ losses from 3DGS~\cite{kerbl3Dgaussians}, while $\mathcal{L}_o$ regularizes vertex opacity~\cite{kheradmand20243d}. The vertex depth alignment loss penalizes the deviation between each vertex depth $z_i$ and the sampled rendered depth $z_i^*$:$\mathcal{L}_{z}=\frac{1}{N}\sum_{i=1}^{N}|z_{i}-z_{i}^{*}$|.
The normal loss $\mathcal{L}_n$ enforces consistency between rendered and surface normals, using supervision from external normal estimators~\cite{yin2023metric3d,hu2024metric3d}, 2DGS self-regularization~\cite{Huang2DGS2024}, or both. The depth loss $\mathcal{L}_d$ further aligns predicted depths with monocular priors from Depth Anything v2~\cite{yang2024depth} after scale-and-shift normalization~\cite{kerbl2024hierarchical}.

In the appearance stage, we freeze mesh connectivity, vertex positions, and opacities, and jointly optimize the residual per-face textures and vertex SH coefficients. The texture objective is defined as
\begin{equation}
\mathcal{L}_{\text{texture}} = \lambda_{l1}\mathcal{L}_{1} + \lambda_{ssim}\mathcal{L}_{\text{SSIM}} + \lambda_{tv}\mathcal{L}_{\text{TV}} + \lambda_{l2}\mathcal{L}_{\text{L2}}.
\end{equation}
The TV loss smooths adjacent texels, while the $\mathcal{L}_2$ term constrains the residual magnitude, ensuring that per-face textures mainly compensate for high-frequency details. Together with real-time Mip-texture pyramids and trilinear sampling, this stage improves multi-scale rendering quality and reduces distant-view aliasing.

\begin{table}[t]
    \centering
    \rmfamily\fontsize{8}{9.6}\selectfont
    \setlength{\tabcolsep}{2.5pt}
    \setlength{\arrayrulewidth}{0.4pt}
    \renewcommand{\arraystretch}{1}

    \begin{minipage}[t]{0.44\textwidth}
        \vspace{0pt}
        \centering
        \caption{\textbf{OpenGL rendering efficiency.}
        All methods are evaluated on the Garden scene using the same
        OpenGL pipeline and rendering configuration at a resolution of
        $1920 \times 1080$ on a MacBook Air M3.
        Storage reports mesh file size on disk.
        OOM denotes an out-of-memory failure.
        Our method achieves the highest frame rate and smallest storage
        footprint.}
        \label{tab:opengl_efficiency}
        \par\vspace{6pt}

        \begin{tabular*}{\linewidth}{@{\extracolsep{\fill}}l|rrr@{}}
            Method & FPS $\uparrow$ & Storage $\downarrow$ & \#Faces $\downarrow$ \\
            \midrule[0.4pt]
            GOF               & OOM & 1.8~GB & 92.1M \\
            RaDe-GS           & OOM & 1.1~GB & 60.8M \\
            MiLo              & 82  & 314~MB & 14.9M \\
            MeshSplatting     & 126 & 152~MB & 7.3M \\
            Gaussian Wrapping & 40  & 638~MB & 31.2M \\
            \midrule[0.4pt]
            Ours & \textbf{236} & \textbf{62~MB} & \textbf{2.7M} \\
        \end{tabular*}
    \end{minipage}%
    \hfill
    \begin{minipage}[t]{0.51\textwidth}
        \vspace{0pt}
        \centering
        \caption{\textbf{Quantitative evaluation under varying mesh counts
        on the Garden scene.}
        Percentages in parentheses represent the ratio of retained mesh
        faces. Compared to MeshSplatting, T$^2$-Splat achieves robust visual
        fidelity even when the mesh size is aggressively reduced
        (e.g., to 10\%).}
        \label{table:mesh_simplification}
        \par\vspace{6pt}

        \begin{tabular*}{\linewidth}{@{\extracolsep{\fill}}l|rrrr@{}}
            Method & PSNR $\uparrow$ & LPIPS $\downarrow$
            & SSIM $\uparrow$ & \#Faces $\downarrow$ \\
            \midrule[0.4pt]
            RaDe-GS           & 22.67 & 0.21 & 0.743 & 60.8M \\
            GOF               & 22.99 & 0.26 & 0.706 & 92.1M \\
            2DGS              & 21.90 & 0.32 & 0.608 & 10.6M \\
            MiLo              & 23.29 & 0.22 & 0.738 & 14.9M \\
            MeshSplatting     & 24.70 & 0.22 & 0.762 & 7.3M \\
            Gaussian Wrapping & 23.45 & 0.27 & 0.717 & 31.2M \\
            \midrule[0.4pt]
            Ours (100\%) & \textbf{25.68} & \textbf{0.15} & \textbf{0.818} & \textbf{2.7M} \\
            Ours (50\%)  & \textbf{25.38} & \textbf{0.17} & \textbf{0.801} & \textbf{1.4M} \\
            Ours (10\%)  & \textbf{23.93} & \textbf{0.30} & \textbf{0.697} & \textbf{0.3M} \\
        \end{tabular*}
    \end{minipage}
\end{table}

\section{Experiments}

In this section, we compare our proposed method with existing approaches on the Mip-NeRF360\cite{barron2022mip}, Tanks \& Temples\cite{knapitsch2017tanks}, and DTU\cite{jensen2014large} datasets. Following MiLo\cite{guedon2025milo} and MeshSplatting\cite{Held2025MeshSplatting}, our evaluation is primarily based on the task of mesh-based novel view synthesis, which measures the performance of reconstructed meshes in complex real-world scenes. Additionally, we calculate the Chamfer Distance (CD) on the DTU dataset to evaluate the accuracy of our mesh reconstruction.

\begin{table}[t]
    \centering
    \caption{\textbf{Ablations on Mip-NeRF 360.}
    We progressively enable Safe RDT, Dynamic Topology Evolution (DTE),
    and residual per-face Mip-textures (MipTex) from baseline (a).
    DTE improves rendering quality with fewer faces, while Mip-textures
    further refine appearance on fixed geometry.
    Results are averaged equally over nine scenes.
    }
    \label{table:mipnerf360_ablation}
    \par\vspace{3pt}
    \normalsize\rmfamily
    \setlength{\tabcolsep}{3pt}
    \renewcommand{\arraystretch}{1}

    \begin{tabular*}{\linewidth}{@{\extracolsep{\fill}}lcccrrrrr@{}}
        \toprule
         & Safe RDT & DTE & MipTex
        & PSNR $\uparrow$
        & LPIPS $\downarrow$
        & SSIM $\uparrow$
        & \#Faces $\downarrow$
        & Time (min) $\downarrow$ \\
        \midrule
        (a) & -- & -- & --
            & 24.66 & 0.30 & 0.734 & 5.3M & 54 \\
        (b) & $\checkmark$ & -- & --
            & 24.81 & 0.31 & 0.740 & 3.0M & 43 \\
        (c) & $\checkmark$ & $\checkmark$ & --
            & 25.12 & 0.29 & 0.756 & \textbf{2.5M} & 57 \\
        (d) & $\checkmark$ & -- & $4\times4$
            & 25.20 & 0.27 & 0.762 & 3.0M & 65 \\
        (e) & $\checkmark$ & -- & $8\times8$
            & 25.31 & 0.24 & 0.769 & 3.0M & 81 \\
        \midrule
        (f) & $\checkmark$ & $\checkmark$ & $4\times4$
            & 25.43 & 0.26 & 0.775 & \textbf{2.5M} & 72 \\
        (g) & $\checkmark$ & $\checkmark$ & $8\times8$
            & \textbf{25.49}
            & \textbf{0.23}
            & \textbf{0.779}
            & \textbf{2.5M}
            & 83 \\
        \bottomrule
    \end{tabular*}
\end{table}

\textbf{Baselines.} We compare our method against 2DGS\cite{Huang2DGS2024}, MiLo\cite{guedon2025milo}, GOF\cite{Yu2024GOF}, RaDe-GS\cite{zhang2026rade}, Gaussian Wrapping\cite{gomez2026blobs}, Triangle Splatting\cite{Held2025Triangle}, and MeshSplatting\cite{Held2025MeshSplatting}. Note that 2DGS, GOF and RaDe-GS require post-processing steps for mesh extraction, whereas MiLo integrates mesh extraction into the optimization process itself. However, all these methods require training a neural color field to texture the mesh (refer to MiLo for detailed procedures). In contrast, Triangle Splatting and MeshSplatting utilize end-to-end mesh training and do not require post-hoc extraction or coloring. Qualitative comparisons demonstrate the results of our method against state-of-the-art baselines, MiLo and MeshSplatting, on the Mip-NeRF360 and Tanks \& Temples datasets.

\textbf{Implementation Details.} All training experiments are conducted on a single NVIDIA RTX 6000 Ada GPU (48GB). Each training run consists of 35,000 iterations, with 30,000 iterations dedicated to geometric optimization and 5,000 iterations for texture refinement. More details are provided in the supplementary material.

\subsection{Mesh-based NVS and Surface Reconstruction}
We report the quantitative results for mesh-based NVS on the Mip-NeRF360 and Tanks \& Temples datasets in Table \ref{tab:comparisons_main_table}. On Mip-NeRF 360, our method significantly outperforms the state-of-the-art MeshSplatting in terms of PSNR, LPIPS, and SSIM, while utilizing fewer vertices and faces. This demonstrates that our method achieves an optimal balance between rendering quality and geometric reconstruction, particularly in capturing high-frequency details. On Tanks \& Temples, our method achieves the best LPIPS and SSIM among the evaluated mesh-based renderings,
while GOF and MiLo obtain higher PSNR. Our method, while outperforming MeshSplatting in PSNR, significantly surpasses all mesh-based baselines in LPIPS and SSIM, proving its superior structural and visual quality, especially in terms of realistic detail recovery. Qualitative results are shown in Figure \ref{fig:visual}.

We report the Chamfer Distance for each method across 15 scenes of the DTU dataset in Table \ref{tab:dtu_sum}. Our method achieves the lowest Chamfer Distance in 3 scenes, which indicates that although our framework is primarily targeted at end-to-end mesh reconstruction and rendering for complex real-world scenes, it still achieves surface reconstruction quality comparable to concurrent methods. Furthermore, our method outperforms MeshSplatting in most scenes, demonstrating that the DTE module—through edge collapse and vertex splitting—effectively improves mesh quality.

\subsection{Rendering Efficiency and Storage}
\label{sec:rendering_efficiency}

Table~\ref{tab:opengl_efficiency} compares OpenGL rendering
efficiency on the Garden scene using a MacBook Air M3 under
the same rendering configuration.
Our method achieves 236 FPS, approximately $1.87\times$ the
frame rate of MeshSplatting (126 FPS), while reducing the mesh
file size from 152 MB to 62 MB, a reduction of approximately
59\%.
GOF and RaDe-GS exceed the available memory under this setup.
These results show that our compact mesh representation
supports efficient rendering with reduced mesh storage
requirements in a standard OpenGL pipeline. For our method, storage and FPS refer to the baked
vertex-colored PLY representation described in the supplementary material.

\subsection{Extreme Mesh Simplification and Robustness Analysis}
As shown in Table \ref{table:mesh_simplification}, we evaluate rendering quality under varying mesh scales on the Garden scene. While existing methods demand massive face counts (e.g., GOF's 92.1M and MiLo's 14.9M) to maintain fidelity, our full model achieves state-of-the-art results across all metrics (PSNR 25.68, LPIPS 0.15) with only 2.7M faces.

To evaluate rendering quality under different mesh budgets,
we control the final face count by adjusting the relative
rates of vertex split and edge collapse in DTE during Stage I.
Each resulting mesh is then held fixed during Stage II,
where residual per-face textures and vertex SH coefficients
are optimized. On Garden, our 50\% configuration achieves higher PSNR,
lower LPIPS, and higher SSIM than MeshSplatting,
using only 1.4M faces compared with 7.3M.
Further reducing the budget to 10\% yields 0.3M faces,
with 23.93~dB PSNR, 0.30 LPIPS, and 0.697 SSIM.
This demonstrates the trade-off between rendering quality
and geometric compactness under increasingly limited
mesh budgets.



\subsection{Ablations}
\label{sec:ablations}
We evaluate the contributions of Safe RDT, Dynamic Topology Evolution
(DTE), and residual per-face Mip-textures on Mip-NeRF 360.
Table~\ref{table:mipnerf360_ablation} reports results averaged over nine scenes.
\paragraph{Safe RDT.}
Comparing configurations (a) and (b), Safe RDT reduces the average
face count from 5.3M to 3.0M, a reduction of approximately 43\%,
while increasing PSNR from 24.66 to 24.81~dB and SSIM from 0.734
to 0.740. LPIPS increases slightly from 0.30 to 0.31.
These results show that constrained RDT face selection provides
a more compact initialization while maintaining comparable
rendering quality.
\paragraph{Dynamic Topology Evolution.}
Adding DTE to Safe RDT, as shown by (b) versus (c), improves PSNR
by 0.31~dB while further reducing the face count from 3.0M to 2.5M.
DTE also improves rendering quality when combined with per-face
textures: comparing (d) with (f) and (e) with (g), it increases
PSNR by 0.23 and 0.18~dB for $4\times4$ and $8\times8$ textures,
respectively. These consistent improvements support the
effectiveness of adapting mesh connectivity through edge collapse
and vertex split operations.
\paragraph{Residual Per-face Mip-textures.}
Appearance refinement improves rendering fidelity on fixed geometry.
Starting from the DTE-refined mesh in (c), adding $4\times4$
textures in (f) increases PSNR from 25.12 to 25.43~dB.
Increasing the texture resolution to $8\times8$ in (g) further
improves PSNR to 25.49~dB, reduces LPIPS from 0.26 to 0.23,
and increases SSIM from 0.775 to 0.779, without changing the
face count. The higher resolution increases training time from
72 to 83 minutes, illustrating the trade-off between appearance
quality and optimization cost.
The complete configuration (g) improves PSNR by 0.83~dB over
baseline (a) and reduces the face count by approximately 53\%,
while increasing total training time from 54 to 83 minutes.

\section{Conclusions}

We presented T$^2$-Splat, a two-stage framework that combines
adaptive mesh topology with residual textures for high-fidelity
rendering. Safe RDT initializes mesh connectivity under local
manifold constraints, while Dynamic Topology Evolution (DTE)
adapts geometric complexity through edge collapses and vertex
splits. After geometry optimization, vertex SH coefficients and
residual per-face Mip-textures are jointly optimized on the fixed
mesh to recover fine appearance details, with scale-adaptive
filtering reducing aliasing. Experiments across multiple datasets
demonstrate improved rendering quality with compact geometry
and competitive surface reconstruction accuracy. Controlling
the relative rates of topology operations further provides
a flexible trade-off between rendering quality and mesh size.
The resulting meshes can be exported in PLY and GLB formats
for use in standard graphics pipelines.
\begin{ack}

We thank the reviewers for their valuable comments and
constructive suggestions, which helped improve this paper.
We also thank the authors and maintainers of the publicly
available datasets and open-source implementations used
in this work for supporting the development and evaluation
of our method. This work is supported by the National Natural Science Foundation of China (NSFC Grant No. 62272184).

\end{ack}


\medskip

{
\small
\bibliographystyle{unsrtnat}
\bibliography{refs.bib}
}


\clearpage
\appendix
\section{Supplementary Material}
\label{sec:supplement}
This supplement expands the implementation and analysis of T$^2$-Splat.
Section~\ref{sec:supp_training} describes the training schedule and parameter groups;
Sections~\ref{sec:supp_safe} and~\ref{sec:supp_dte} detail mesh initialization and topology updates;
Section~\ref{sec:supp_texture} explains residual texture coordinates and filtering;
Section~\ref{sec:supp_export} distinguishes the trained representation from exported assets;
and Sections~\ref{sec:supp_results} and~\ref{sec:supp_limits} discuss additional results and limitations.

\subsection{Training Schedule and Configuration}
\label{sec:supp_training}
All training experiments use a single NVIDIA RTX 6000 Ada GPU with 48\,GB of memory.
Each run contains 35,000 iterations: 30,000 iterations for geometry and base
appearance optimization, followed by 5,000 iterations of appearance refinement.
Table~\ref{tab:hyperparameters} lists the reported hyperparameters, and
Table~\ref{tab:supp_schedule} separates the roles of the different training intervals.

\begin{table}[ht]
\centering
\caption{\small Main hyperparameters used in our experiments.}
\begin{tabular}{l|cc}

\toprule

 & Outdoor & Indoor \\
\midrule
feature learning rate & 0.0016  & 0.004 \\
opacity learning rate & 0.03 & 0.05 \\
vertex pos. learning rate & 0.0015 & 0.0015 \\
texture learning rate & 0.0005 & 0.0005 \\
\midrule
densification start & 500  & 500 \\
densification end & 10000 & 10000 \\
densification interval & 500 & 500 \\
start pruning & 4000 & 3000 \\
pruning threshold & 0.235 & 0.235 \\
mesh creation & 11000 & 11000 \\
DTE start & 12000 & 12000 \\
DTE end & 20000 & 20000 \\
\midrule
$\beta_z$ & 0.00025 & 0.00025 \\
$\beta_n$ & 0.0001 & 0.0001 \\
$\beta_d$ & 0.01 & 0.0 \\
$\beta_o$ & 2e-06 & 0.0 \\
$\lambda_{l1}$ & 0.8 & 0.8 \\
$\lambda_{ssim}$ & 0.2 & 0.2 \\
$\lambda_{tv}$ & 0.01 & 0.01 \\
$\lambda_{l2}$ & 0.01 & 0.01 \\

\bottomrule
\end{tabular}

\label{tab:hyperparameters}
\end{table}

\begin{table}[htbp]
\centering
\caption{Training stages and their roles. Iteration indices describe the reported
schedule; the beginning and end of each interval mark stage transitions.}
\label{tab:supp_schedule}
\small
\setlength{\tabcolsep}{5pt}
\renewcommand{\arraystretch}{1.12}
\begin{tabular}{@{}>{\raggedright\arraybackslash}p{0.20\linewidth}>{\raggedright\arraybackslash}p{0.27\linewidth}>{\raggedright\arraybackslash}p{0.46\linewidth}@{}}
\toprule
Iteration / event & Representation update & Optimization role \\
\midrule
500--10,000 & Densification every 500 iterations & Establish geometric support before mesh construction; pruning starts at 4,000 outdoors and 3,000 indoors. \\
11,000 & Safe RDT & Replace soup connectivity with a constrained face set; retain vertex parameters and optimizer states. \\
12,000--20,000 & DTE every 1,000 iterations & Adapt connectivity using edge collapse and gradient-driven refinement while optimizing geometry and base appearance. \\
20,000--30,000 & DTE interval completed & Continue Stage I optimization of vertex positions, opacities, and SH coefficients. \\
30,000--35,000 & Fixed topology, positions, opacity & Jointly optimize vertex SH coefficients and residual per-face textures. \\
\bottomrule
\end{tabular}
\end{table}

\paragraph{Topology update budget.}
Within the DTE interval, edge collapse targets 5\%--10\% of the current vertices,
whereas refinement targets the top 5\% ranked by perspective-compensated
position gradients. Local validity constraints determine which candidate edits are accepted.
The resulting face-count change depends on the local neighborhood and whether
the edited edge is on a boundary.
The separate 50\% and 10\% configurations in
Table~\ref{table:mesh_simplification} refer to retained face budgets relative to
the full Garden configuration, obtained by adjusting the relative rates of the two operations.

\paragraph{Appearance-stage parameter groups.}
Residual textures use $8\times8$ RGB grids by default and are initialized to zero.
Their base learning rate is $5\times10^{-4}$ with exponential decay; the reported
learning rate for higher-order SH coefficients during refinement is
$2\times10^{-4}$. Geometry and opacity parameters remain fixed in this stage.
The $4\times4$ variant in Table~\ref{table:mipnerf360_ablation} changes texture
resolution, with geometry held fixed within each appearance-refinement run.
The appearance objective uses weights $0.8$, $0.2$, $0.01$, and $0.01$ for
the photometric $\mathcal{L}_1$, SSIM-based, total-variation, and residual-magnitude
terms, respectively. The latter two terms discourage oscillatory texels and
unnecessarily large residual corrections.

\subsection{Safe RDT: Candidate Selection and Local Connectivity}
\label{sec:supp_safe}
 To construct a surface mesh from the optimized triangle soup,
we extend the Restricted Delaunay Triangulation (RDT) procedure
used in MeshSplatting~\cite{Held2025MeshSplatting}.
We first perform Delaunay tetrahedralization on the optimized
vertex set using SciPy and construct dual segments by connecting
the circumcenters of adjacent tetrahedra. A Bounding Volume
Hierarchy (BVH) accelerates intersection queries against the
input triangle soup, with M\"oller--Trumbore tests performed
at the leaf nodes. BVH traversal and intersection testing
are accelerated using Taichi. Internal tetrahedral faces
whose dual segments intersect the input surface form the
candidate face set.

We further introduce \emph{Safe RDT} to filter these candidates
and select a consistently oriented surface.
For each candidate face $f$, let $\mathbf{c}_f$,
$\mathbf{n}_f$, and $A_f$ denote its centroid, unit normal,
and area. Let $\mathbf{p}_f$ be the closest point on the
nondegenerate triangles of the input soup, and let $g_f$
denote the corresponding input triangle. We evaluate
\begin{equation}
\begin{aligned}
d_f &=
\frac{\|\mathbf{c}_f-\mathbf{p}_f\|_2}
{\max(\ell_{\max}(g_f),\epsilon)}, \\
a_f &= \left|\mathbf{n}_f^\top\mathbf{n}_{g_f}\right|, \\
q_f &=
\frac{4\sqrt{3}A_f}
{\sum_{e\in E(f)}\ell_e^2},
\end{aligned}
\label{eq:safe_rdt_criteria}
\end{equation}
where $\ell_{\max}(g_f)$ is the longest edge of $g_f$,
and $\ell_e$ is the length of an edge in $E(f)$.
These quantities measure normalized surface distance,
unsigned normal agreement, and triangle quality, respectively.
We retain nondegenerate candidates satisfying
$d_f\leq0.5$, $a_f\geq0.15$, and $q_f\geq0.02$,
and rank them using
\begin{equation}
s_f=a_f\exp(-4d_f)\frac{1+q_f}{2}.
\label{eq:safe_rdt_ranking}
\end{equation}

Starting from an empty face set, we greedily grow the surface
through a score-prioritized edge frontier. Candidate insertion
must introduce no duplicate face, keep each edge incident to
at most two faces, preserve a single path or cycle in the
link of every used vertex, and admit consistent face orientations.
Candidates that temporarily violate the vertex-fan condition
are deferred and reconsidered as neighboring faces are accepted.
A new component is seeded when the frontier is exhausted.
Orientation consistency is maintained using a disjoint-set
structure with parity information.

Safe RDT updates only face connectivity and winding, retaining
vertex positions, opacity and SH parameters, and their optimizer
states. It permits open boundaries and multiple connected components.
Geometric filtering favors local fidelity, while the topology constraints
maintain valid surface neighborhoods.
The selected mesh initializes DTE, which further adapts
its connectivity through edge collapse and vertex split
operations.

\paragraph{Interpretation of the geometric criteria.}
The distance score $d_f$ is dimensionless: normalizing by the local longest edge
compares displacement relative to the support scale of the input soup.
The absolute dot product $a_f$ measures normal agreement independently of
input winding, while the connectivity-selection procedure determines
consistent face orientations.
For a nondegenerate triangle, $q_f$ lies in $(0,1]$, equals one for an equilateral
triangle, and tends to zero as the triangle degenerates. It is invariant to
uniform scaling because both area and squared edge length scale quadratically.
Thus the three tests penalize different failure modes: distance from the input
surface, incompatible local normals, and poorly shaped triangles.

\paragraph{Joint edge and vertex constraints.}
Edge-incidence constraints limit each edge to at most two incident faces.
Vertex-link constraints additionally require a single path or cycle around
each used vertex, corresponding to boundary or interior surface neighborhoods.
Together, these conditions exclude disconnected face fans at a shared vertex.
Deferral allows a temporarily disconnected candidate fan to become connected
as neighboring faces are accepted during greedy surface growth.

\subsection{Dynamic Topology Evolution and Optimizer Continuity}
\label{sec:supp_dte}

\paragraph{QEM minimization in Cartesian coordinates.}
For an edge $(u,v)$, write the merged quadric in block form as
\begin{equation}
Q_{uv}=\begin{bmatrix}H&\mathbf{b}\\\mathbf{b}^{\top}&c\end{bmatrix},
\qquad
E(\mathbf{x})=\mathbf{x}^{\top}H\mathbf{x}
               +2\mathbf{b}^{\top}\mathbf{x}+c.
\label{eq:supp_qem_block}
\end{equation}
Differentiating gives $\nabla E(\mathbf{x})=2H\mathbf{x}+2\mathbf{b}$.
When $H$ is invertible, the unconstrained minimizer satisfies
$H\mathbf{x}^{*}=-\mathbf{b}$. For a singular system, the endpoint and midpoint
fallback in the main paper selects the lowest-error member of
$\{\mathbf{u},\mathbf{v},(\mathbf{u}+\mathbf{v})/2\}$.
The minimizer supplies a candidate location, which is evaluated using
the appearance-aware cost and validity checks.

\paragraph{Boundary protection as an additional quadric.}
The standard boundary construction uses a plane through the edge and perpendicular to its incident face. For an open boundary edge, let $\mathbf{t}$ be its unit tangent and
$\mathbf{n}_f$ the normal of its incident face. A plane containing the edge
and perpendicular to that face has unit normal
\begin{equation}
\mathbf{n}_b=\frac{\mathbf{t}\times\mathbf{n}_f}
{\|\mathbf{t}\times\mathbf{n}_f\|_2},
\qquad
\mathbf{p}_b=
\begin{bmatrix}\mathbf{n}_b\\-\mathbf{n}_b^{\top}\mathbf{u}\end{bmatrix}.
\label{eq:supp_boundary_plane}
\end{equation}
With the reported boundary weight of 10, this construction contributes
$10\,\mathbf{p}_b\mathbf{p}_b^{\top}$ to the endpoint quadrics.
Each endpoint contribution to the error is
$10[\mathbf{n}_b^{\top}(\mathbf{x}-\mathbf{u})]^2$.
Together with the incident surface plane, it discourages movement away from
the boundary line while permitting movement along it. This soft geometric
penalty is combined with topology and orientation checks when accepting collapses.

\paragraph{Admissibility and simultaneous edits.}
The full simplicial link condition in Eq.~\eqref{eq:link_condition} includes
both vertices and edges in the links. Boundary, merged-fan, degeneracy,
and orientation tests further screen candidate changes for local validity.
Within one update, accepted collapses have disjoint affected one-rings, which
prevents two selected collapses from simultaneously modifying the same local
connectivity.

\paragraph{Refinement and a conforming edge subdivision.}
The refinement score multiplies a vertex-position gradient magnitude by its
camera-space depth and averages across the contributing views. This score
prioritizes vertices whose positions remain sensitive to the image-space objective.
For a selected edge $(i,j)$, the new midpoint and its initial appearance
attributes follow the interpolation formulas in the main paper.
For the subdivision to remain conforming, the same midpoint must be shared
by all incident faces of that edge: an interior edge has two such faces and
a boundary edge has one. Each incident triangle is replaced by two triangles
with consistent winding. Sharing the midpoint across incident faces
avoids T-junctions in the refined neighborhood.

\paragraph{Parameter interpolation and optimizer-state interpolation.}
The attribute and moment-transfer rules serve different purposes. Collapsed
appearance attributes use opacity-weighted averaging, whereas Adam moments
use the incident-area weights in Eq.~\eqref{eq:collapse_optimizer_state}.
Writing $\eta=A_u/(A_u+A_v)$ for positive total incident area gives
$m_{\mathrm{new}}=\eta m_u+(1-\eta)m_v$ and
$v_{\mathrm{new}}=\eta v_u+(1-\eta)v_v$.
The latter remains elementwise nonnegative because it is a convex combination
of second-moment estimates. For splitting, the two perturbed first moments in
Eq.~\eqref{eq:split_optimizer_state} satisfy
\begin{equation}
\frac{m_k^{+}+m_{k'}}{2}=m_k,\qquad
v_k^{+}=v_{k'}=v_k.
\label{eq:supp_moment_identity}
\end{equation}
The opposite perturbations therefore preserve the pair's mean first moment
while breaking identical inherited states. These transfers initialize the
updated parameterization from the accumulated optimization history.

\subsection{Residual Texture Coordinates, Filtering, and Capacity}
\label{sec:supp_texture}

\paragraph{A continuous transition to appearance refinement.}
The rendered color combines the SH-based appearance with a spatial residual.
At initialization, $T_f=0$ for every face, and average pooling and linear
texture interpolation both map zero to zero. Consequently,
$C_{\mathrm{res}}^{\lambda}=0$ at every mip level, so adding the residual branch
initially preserves the base rendering. During refinement, SH coefficients
and textures are optimized jointly. The residual models spatial variation
in local face coordinates, while the SH coefficients provide view-dependent appearance.

\paragraph{Triangular support within a square texture.}
The barycentric coordinates $(u,v)=(w_j,w_k)$ belong to
$\mathcal{D}=\{(u,v):u\geq0,v\geq0,u+v\leq1\}$.
The mirror extension can be written explicitly as
\begin{equation}
\mu(u,v)=
\begin{cases}
(u,v), & u+v\leq1,\\
(1-v,1-u), & u+v>1.
\end{cases}
\label{eq:supp_mirror}
\end{equation}
Reflection fixes the diagonal $u+v=1$ and maps the complementary half of the
unit square back into $\mathcal{D}$. It provides values outside the triangle
for square-grid interpolation and filtering within each face.

\paragraph{Screen-space footprint and coordinate units.}
The LOD estimate is most explicit when its derivatives are expressed in texel
units. For an $R\times R$ base texture and normalized $(u,v)$, define
\begin{equation}
\begin{aligned}
\rho_x &= R\sqrt{(\partial u/\partial x)^2+(\partial v/\partial x)^2},\\
\rho_y &= R\sqrt{(\partial u/\partial y)^2+(\partial v/\partial y)^2},\\
\lambda_{\mathrm{tex}} &=
\log_2\!\left(\max(\rho_x,\rho_y)+\epsilon\right).
\end{aligned}
\label{eq:supp_texel_lod}
\end{equation}
Here $x,y$ are pixel coordinates and $\rho=\max(\rho_x,\rho_y)$ measures texels covered per pixel.
The factor $R$ is needed for normalized UVs; it is already absorbed when
derivatives are taken in texel coordinates. Equation~\eqref{eq:supp_texel_lod} makes the normalized-coordinate formulation explicit, following the texel-space footprint convention in standard texture filtering.
For mip levels $0,\ldots,L$, valid level selection uses
\begin{equation}
\hat\lambda=\min\{L,\max\{0,\lambda_{\mathrm{tex}}\}\},\quad
l_0=\lfloor\hat\lambda\rfloor,\quad
l_1=\min(l_0+1,L),\quad
\delta=\hat\lambda-l_0.
\label{eq:supp_lod_clamp}
\end{equation}
The filtered residual is then
\begin{equation}
C_{\mathrm{res}}=(1-\delta)\operatorname{Bilinear}(T_f^{l_0},u,v)
+\delta\operatorname{Bilinear}(T_f^{l_1},u,v).
\label{eq:supp_trilinear}
\end{equation}
This form makes the finest- and coarsest-level boundary cases explicit.
Because coarse levels are produced by average pooling, their gradients
propagate to the underlying fine-level texture. Isotropic mip filtering
reduces high-frequency texture content under minification.

\paragraph{Appearance capacity and storage scaling.}
For $F$ faces, the dense base-level RGB grids contain $3FR^2$ scalar entries.
When $R=2^L$, storing every mip level requires
\begin{equation}
3F\sum_{l=0}^{L}(R/2^l)^2
=3F\frac{4R^2-1}{3}<4FR^2
\label{eq:supp_mip_storage}
\end{equation}
scalar entries. For $R=8$, a full pyramid has $64+16+4+1=85$ RGB texels
per face, compared with 64 at its base level. Memory use also depends on
precision, gradients, optimizer states, and the storage of intermediate mip levels.
Increasing $R$ from 4 to 8 quadruples the base-grid capacity at a fixed face
count. Conversely, reducing the number of faces also reduces the total number
of face textures, even when each remaining face has the same resolution.

\subsection{Mesh Export and Rendering Efficiency}
\label{sec:supp_export}
Both PLY and GLB export first aggregate SH colors, including higher-order
contributions, over visible training views and incorporate residual textures
to produce static baked appearance. Both formats preserve the face count.

\paragraph{Vertex-colored PLY.}
Residual textures are sampled and fitted to three corner-color residuals per
face. The corner residuals are combined at shared vertices using face-area
weights and added to the baked SH colors. The file stores positions, RGB
vertex colors, and face indices, preserving the geometry and connectivity.
The baked vertex colors consolidate the SH and residual texture contributions
into a single vertex-color representation.

\paragraph{Texture-atlas GLB.}
Interpolated SH colors and residual textures are combined into per-face RGB
textures and packed into an atlas. The GLB contains geometry, UV coordinates,
embedded PNG textures, and materials. The face count is unchanged, while
vertices are expanded into three corners per face to accommodate independent
UV coordinates. This retains more intra-face appearance capacity than vertex
color fitting, at the cost of additional storage.

\paragraph{OpenGL rendering efficiency.}
Table~\ref{tab:opengl_efficiency} evaluates the baked vertex-colored PLY on
Garden at $1920\times1080$ on a MacBook Air M3, reporting 236\,FPS and a
file size of 62\,MB.
Relative to MeshSplatting under that OpenGL setup, the displayed values imply
$1.87$ times the frame rate and a
$59.2\%$ reduction in file size.

\subsection{Additional Results and Analysis}
\label{sec:supp_results}

\subsubsection{Surface reconstruction and topology}
\paragraph{Geometric reconstruction on Tanks \& Temples.}
Table~\ref{tab:tnt_geometry} reports F-scores on six scenes using
scale-corrected virtual scanning. T$^2$-Splat achieves an arithmetic mean
of 0.40, computed from unrounded per-scene values, and the highest
Courthouse score of 0.26 among the listed methods. In the full-scene
columns, MiLo and Gaussian Wrapping obtain means of 0.42 and 0.53,
respectively. The table groups foreground-only and full-scene extraction
results separately. These surface measurements complement the NVS evaluation.

\begin{table}[t]
    \centering
    \caption{\textbf{Geometric reconstruction on Tanks \& Temples.}
    F-scores ($\uparrow$) on six scenes and their arithmetic mean.
    The mean for T$^2$-Splat is computed from unrounded
    per-scene scores.}
    \label{tab:tnt_geometry}

    \begingroup
    \footnotesize
    \setlength{\tabcolsep}{4pt}
    \renewcommand{\arraystretch}{1.1}
    \resizebox{\linewidth}{!}{%
        \begin{tabular}{@{}lcc|ccccc@{}}
            \toprule
            & \multicolumn{2}{c|}{Foreground only}
            & \multicolumn{5}{c}{Full scene extraction} \\
            \cmidrule(lr){2-3}
            \cmidrule(l){4-8}
            Scene
                & 2DGS
                & Gaussian Wrapping
                & GOF
                & RaDe-GS
                & MiLo
                & Gaussian Wrapping
                & \textbf{T$^2$-Splat} \\
            \midrule
            Barn
                & 0.51 & 0.67 & 0.49 & 0.47 & 0.52 & 0.67 & 0.47 \\
            Caterpillar
                & 0.31 & 0.50 & 0.29 & 0.30 & 0.33 & 0.50 & 0.26 \\
            Courthouse
                & 0.10 & 0.16 & 0.11 & 0.11 & 0.12 & 0.18 & 0.26 \\
            Ignatius
                & 0.61 & 0.78 & 0.63 & 0.66 & 0.74 & 0.78 & 0.66 \\
            Meetingroom
                & 0.19 & 0.41 & 0.24 & 0.25 & 0.23 & 0.41 & 0.21 \\
            Truck
                & 0.54 & 0.63 & 0.50 & 0.57 & 0.56 & 0.63 & 0.50 \\
            \midrule
            Mean
                & 0.38 & 0.53 & 0.38 & 0.39 & 0.42 & 0.53 & 0.40 \\
            \bottomrule
        \end{tabular}%
    }
    \endgroup

    \par\smallskip
\end{table}

\begin{table}[t]
\centering
\caption{
Geometry and topology statistics of the final meshes on the
nine scenes of Mip-NeRF 360.
Triangle shape quality is reported as the equally weighted
average of the per-scene median values.}
\label{tab:geometry_ablation}
\small
\setlength{\tabcolsep}{6pt}
\renewcommand{\arraystretch}{1.15}
\resizebox{\linewidth}{!}{
\begin{tabular}{lccccc}
\toprule

& \shortstack{Average faces\\(M)}
& \shortstack{Non-manifold\\edges (\%) $\downarrow$}
& \shortstack{Non-manifold\\vertices (\%) $\downarrow$}
& \shortstack{Triangle shape\\quality $\uparrow$}
& \shortstack{Faces in largest\\component (\%)} \\
\midrule
A: Baseline
& 5.308
& 17.84
& 65.06
& 0.5796
& 52.94 \\

B: Safe RDT
& 3.049
& \textbf{0}
& \textbf{0}
& 0.5918
& 44.30 \\

C: Safe RDT + DTE
& \textbf{2.468}
& \textbf{0}
& \textbf{0}
& \textbf{0.6203}
& \textbf{54.41} \\
\bottomrule
\end{tabular}
}
\end{table}

\paragraph{Complementary evidence from DTU.}
Table~\ref{tab:dtu_sum} reports lower Chamfer distance than MeshSplatting
on 14 of the 15 DTU scans; scan 97 is the exception (1.18 versus 1.15).
The reported mean decreases from 0.89 to 0.68, a reduction of approximately
23.6\%. T$^2$-Splat achieves the lowest reported
error on scans 63, 83, and 105. Its mean is tied with RaDe-GS and MiLo, while Gaussian Wrapping has a lower mean of 0.49.

\paragraph{Compactness and local mesh quality.}
Table~\ref{tab:geometry_ablation} reports geometry statistics on nine Mip-NeRF 360
scenes. Safe RDT reduces the average face count from 5.308M to 3.049M
($42.6\%$). DTE further reduces it to 2.468M ($19.1\%$ relative to Safe RDT),
for an overall reduction of $53.5\%$ relative to the baseline.
Both constrained configurations have zero measured non-manifold edge and
vertex ratios in these evaluations.
The reported triangle-quality statistic is the equally weighted average
of the per-scene median values.
DTE increases this statistic from 0.5918 to 0.6203, an absolute gain of 0.0285.

\paragraph{Component connectivity.}
Safe RDT decreases the fraction of faces in the largest component from
52.94\% to 44.30\%, while reducing the local non-manifold ratios to zero.
Adding DTE raises this fraction to 54.41\%, a gain of 10.11 percentage
points over Safe RDT. Together with the reduced face count and improved
triangle quality, this shows how DTE changes both mesh compactness and
the distribution of faces across connected components.

\subsubsection{Component contributions and resource trade-offs}
\paragraph{Controlled comparisons in the ablation table.}
Table~\ref{tab:supp_ablation_deltas} summarizes selected differences from
Table~\ref{table:mipnerf360_ablation}. Comparing rows with the same
texture resolution shows the effect of DTE, while rows with the same
geometry configuration show the effect of appearance refinement.
The appearance-refinement stage jointly optimizes SH coefficients and textures.

\begin{table}[htbp]
\centering
\caption{Differences derived from the displayed values in
Table~\ref{table:mipnerf360_ablation}. Positive PSNR/SSIM changes and negative
LPIPS changes indicate improvements. Time differences refer to reported
total training time.}
\label{tab:supp_ablation_deltas}
\small
\setlength{\tabcolsep}{4pt}
\begin{tabular*}{\linewidth}{@{\extracolsep{\fill}}lrrrr@{}}
\toprule
Comparison & $\Delta$PSNR (dB) & $\Delta$LPIPS & $\Delta$SSIM & $\Delta$time (min) \\
\midrule
(b)$\to$(c): add DTE & +0.31 & -0.02 & +0.016 & +14 \\
(e)$\to$(g): add DTE with $8^2$ textures & +0.18 & -0.01 & +0.010 & +2 \\
(c)$\to$(f): add $4^2$ refinement & +0.31 & -0.03 & +0.019 & +15 \\
(f)$\to$(g): increase $4^2$ to $8^2$ & +0.06 & -0.03 & +0.004 & +11 \\
(a)$\to$(g): complete configuration & +0.83 & -0.07 & +0.045 & +29 \\
\bottomrule
\end{tabular*}
\end{table}

With DTE enabled, increasing texture resolution from $4\times4$ to
$8\times8$ quadruples base texture capacity, improves average PSNR by
0.06\,dB, and reduces LPIPS by 0.03. Total training time rises from 72 to
83 minutes, illustrating the quality--cost trade-off of higher texture
resolution. Configurations (e) and (g) differ by 2 minutes in total training
time and also use different face counts.

\paragraph{Mesh budgets on Garden.}
At the 50\% face-budget setting, PSNR decreases by 0.30\,dB, LPIPS increases
by 0.02, and SSIM decreases by 0.017 relative to the full model.
The resulting 1.4M-face mesh still improves over the listed MeshSplatting
result by 0.68\,dB PSNR, 0.05 LPIPS, and 0.039 SSIM while using approximately
$80.8\%$ fewer faces than its 7.3M-face mesh.
At the 10\% setting, the corresponding changes relative to the full model
are $-1.75$\,dB PSNR, $+0.15$ LPIPS, and $-0.121$ SSIM.
These settings provide a tunable balance between rendering quality and
mesh size.

\subsubsection{Per-scene novel view synthesis}
Tables~\ref{tab:tandt}, \ref{tab:mipnerf_outdoor}, and~\ref{tab:mipnerf_indoor}
report per-scene measurements for the two Tanks \& Temples NVS scenes
and all nine Mip-NeRF 360 scenes. The two-scene NVS evaluation uses a different protocol and scene set
from the six-scene geometric evaluation above; Truck appears in both.
Among the outdoor Mip-NeRF 360 scenes, Garden has the lowest LPIPS (0.148)
and highest SSIM (0.818), whereas Flowers has the lowest PSNR (19.81\,dB)
and SSIM (0.546). Indoor PSNR values range from 27.01\,dB on Counter to
29.36\,dB on Bonsai. For Tanks \& Temples, Truck exceeds Train by 3.59\,dB PSNR and 0.083 SSIM,
and has 0.092 lower LPIPS.

\begin{table}[ht]
    \centering
    \caption{Per scene LPIPS, PSNR, and SSIM scores for the Truck and Train scenes of the T\&T dataset.}    
    \tabcolsep=0.1cm
    \resizebox{0.45\columnwidth}{!}{      
    \begin{tabular}{l|cc}
     & Truck & Train \\
    \midrule
    PSNR$\uparrow$ & 22.91 & 19.32 \\
    LPIPS$\downarrow$ & 0.167 & 0.259 \\
    SSIM$\uparrow$ & 0.836 & 0.753 \\
    \end{tabular}
    }    
    \footnotesize

    \label{tab:tandt}
\end{table}

\begin{table}[ht]
    \centering
    \caption{Per-scene PSNR, LPIPS, and SSIM scores for outdoor MipNeRF360 scenes.}
    \tabcolsep=0.1cm
    \resizebox{0.98\columnwidth}{!}{      
    \begin{tabular}{l|ccccc}
         & Bicycle & Flowers & Garden & Stump & Treehill \\
        \midrule
        PSNR$\uparrow$  & 23.96 & 19.81 & 25.68 & 25.36 & 21.01 \\
        LPIPS$\downarrow$ & 0.247 & 0.336 & 0.148 & 0.252 & 0.327 \\
        SSIM$\uparrow$  & 0.709 & 0.546 & 0.818 & 0.723 & 0.595 \\
    \end{tabular}
    }

    \label{tab:mipnerf_outdoor}
\end{table}
\begin{table}[ht]
    \centering
    \caption{Per-scene PSNR, LPIPS, and SSIM scores for indoor MipNeRF360 scenes.}
    \tabcolsep=0.1cm
    \resizebox{0.8\columnwidth}{!}{      
    \begin{tabular}{l|cccc}
         & Room & Counter & Kitchen & Bonsai \\
        \midrule
        PSNR$\uparrow$  & 28.79 & 27.01 & 28.45 & 29.36 \\
        LPIPS$\downarrow$ & 0.224 & 0.214 & 0.150 & 0.214 \\
        SSIM$\uparrow$  & 0.904 & 0.885 & 0.909 & 0.924 \\
    \end{tabular}
    }

    \label{tab:mipnerf_indoor}
\end{table}

\begin{figure}[htbp]
\centering
\includegraphics[width=\linewidth]{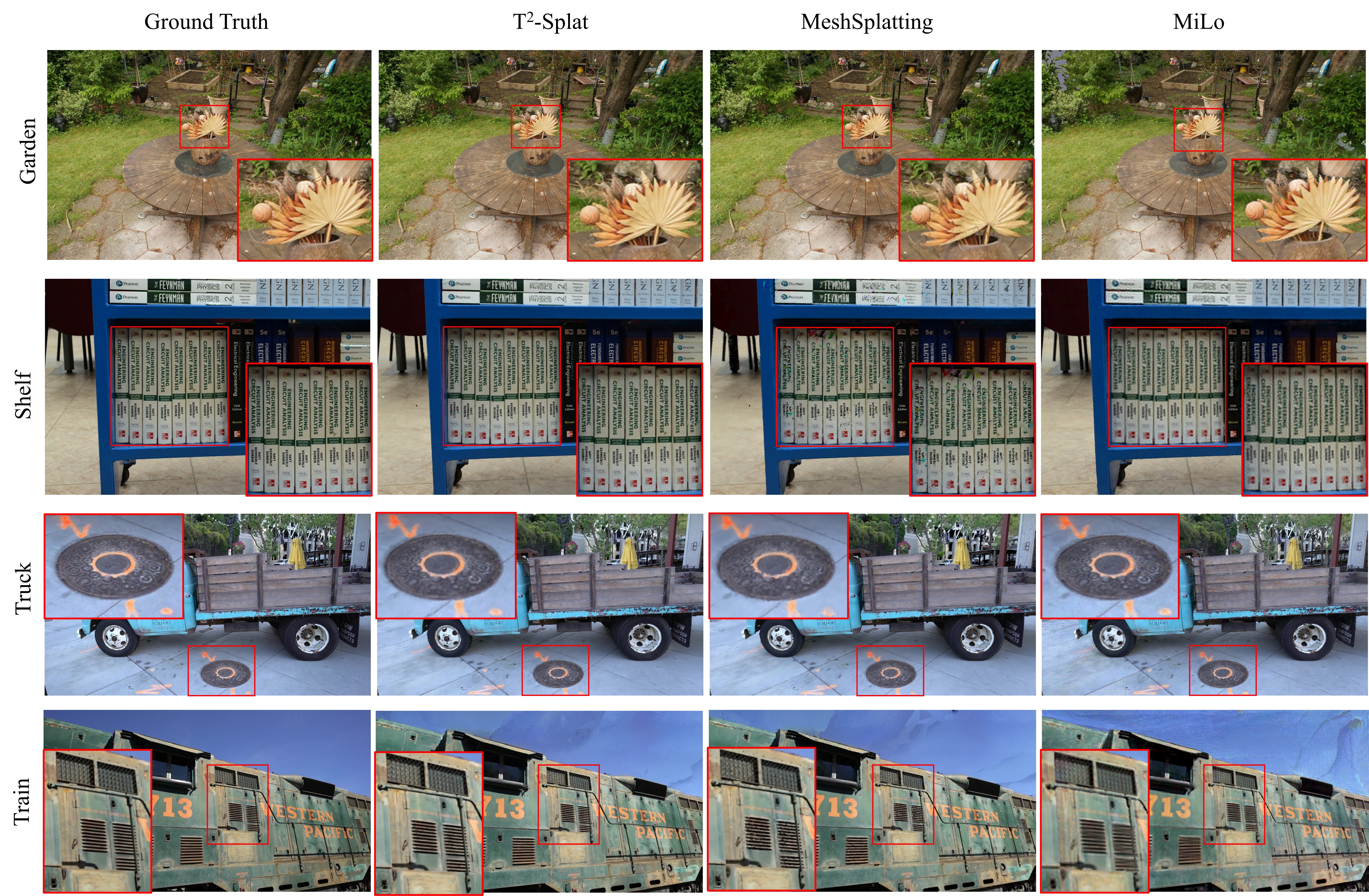}
\caption{Additional qualitative comparisons on Garden, Shelf, Truck, and
Train. The columns show ground truth, T$^2$-Splat, MeshSplatting, and MiLo.
The highlighted regions provide close views of local texture and edge detail.
Shelf is an additional STRinGS-360 example, outside the NVS averages
reported in the per-scene tables.}
\label{fig:visual1}
\end{figure}

\paragraph{Qualitative comparisons.}
Figure~\ref{fig:visual1} shows Garden, Shelf, Truck, and Train, including
enlarged image regions. These examples allow visual comparison of local
texture patterns and edge detail against ground truth, MeshSplatting, and MiLo.
Shelf is an additional STRinGS-360 example and is not part of the nine-scene
Mip-NeRF 360 or two-scene Tanks \& Temples NVS averages.

\subsection{Limitations and Future Directions}
\label{sec:supp_limits}

\paragraph{Surface smoothness and geometric completeness.}
The reconstructed mesh still exhibits local surface undulations and normal
discontinuities, while the recovery of thin structures and sparsely observed
regions also depends on the initial geometric support. Future work could
jointly train the explicit mesh with a signed distance field
(SDF), combining continuous geometric constraints with mesh
rendering and topology optimization. Surface-position and normal consistency
could couple the two representations. Together with multi-view observations and
moderate smoothing regularization, this approach could improve surface
smoothness and geometric coherence while preserving sharp edges and thin structures.

\paragraph{Local topology updates and global geometry.}
Connectivity checks evaluate the neighborhoods affected by an edit, whereas
surface self-intersections depend on the spatial arrangement of faces that
need not share a local neighborhood. Continuous vertex optimization can also fold a locally valid
mesh. Efficient collision queries and displacement constraints could therefore
complement the topology checks during both editing and subsequent optimization.
The current discrete editing procedure additionally involves edge extraction,
candidate ranking, and transfers between CPU and GPU memory. Incremental
adjacency maintenance, local candidate updates, and parallel optimizer-state
transfer offer directions for reducing this overhead on larger scenes.

\paragraph{Geometry and texture resource allocation.}
Dense per-face RGB textures contain $3FR^2$ scalar parameters for $F$ faces and
resolution $R$, with additional storage for gradients, optimizer states, and
mip levels. A uniform resolution assigns the same capacity to faces with very
different projected sizes and appearance complexity. Reducing the face count
also reduces the number of texture grids, so larger retained triangles must
represent more surface detail within each grid. Joint allocation of face and
texel budgets could account for projected area, image residuals, and view
coverage. Adaptive texture resolution, shared texture bases, and quantization
could further improve storage efficiency under a specified quality budget.

\paragraph{Texture continuity and oblique views.}
Independent face textures can assign different colors to the same shared edge.
The mirror extension provides samples outside each triangular domain, but does
not coordinate the values of neighboring faces. Cross-face consistency terms,
shared edge samples, or jointly optimized atlas boundaries could improve
continuity. Filtering presents a related challenge: isotropic mip levels use
a single scale for footprints that can become strongly elongated at grazing
angles. Differentiable anisotropic filtering based on the projected texture
footprint could better preserve detail in the short-footprint direction while
suppressing aliasing along the long direction.

\paragraph{Complex appearance and lighting variation.}
Vertex SH coefficients represent directional appearance, while residual
textures describe spatial variation in fixed face coordinates. Sharp specular
highlights combine both types of variation and can demand more capacity than
this decomposition provides. Transparency and refraction additionally involve
layered visibility and light transport through surfaces. Direction-conditioned
residuals, local material models, and layered surface representations are
possible extensions for these effects. The learned appearance also incorporates
the illumination of the capture environment. Separating material properties
from lighting would enable editing and relighting beyond the captured conditions.

\paragraph{Export fidelity and deployment.}
Static color baking compresses the trained appearance for conventional mesh
rendering. It removes directional variation, and fitting a face texture to
three corner colors can smooth spatial detail within the face. Texture-atlas
export retains more of this variation but increases texture and UV storage.
Future export methods could optimize the representation directly for the
target renderer and a chosen asset-size or frame-time budget. Combining mesh
levels of detail with texture resolution selection and atlas packing could
support scalable deployment; a lightweight directional shader could retain
part of the SH appearance when the rendering environment permits it.

\clearpage
\begingroup
\small
\section*{NeurIPS Paper Checklist}

\begin{enumerate}

\item {\bf Claims}
    \item[] Question: Do the main claims made in the abstract and introduction accurately reflect the paper's contributions and scope?
    \item[] Answer: \answerYes{} 
    \item[] Justification: We introduce T$^2$-Splat, a unified differentiable rendering framework that bridges the gap between high-fidelity 3D Gaussian Splatting  and industry-standard polygon pipelines. By integrating dynamic topology evolution with scale-adaptive texturing, our method achieves superior rendering quality and surface reconstruction with a fraction of the geometric complexity of existing baselines. Please see our abstract and introduction for further details.

\item {\bf Limitations}
    \item[] Question: Does the paper discuss the limitations of the work performed by the authors?
    \item[] Answer: \answerYes{} 
    \item[] Justification: Our method still struggles with the geometric crumpling and transparent objects like most mesh-based reconstruction methods. Please see the appendix for further details.

\item {\bf Theory assumptions and proofs}
    \item[] Question: For each theoretical result, does the paper provide the full set of assumptions and a complete (and correct) proof?
    \item[] Answer: \answerYes{} 
    \item[] Justification: In the method and experiment sections, we provide a detailed discussion to demonstrate each result.

    \item {\bf Experimental result reproducibility}
    \item[] Question: Does the paper fully disclose all the information needed to reproduce the main experimental results of the paper to the extent that it affects the main claims and/or conclusions of the paper (regardless of whether the code and data are provided or not)?
    \item[] Answer: \answerYes{} 
    \item[] Justification: We have disclosed the core technical information, implementation details, and evaluation methods required to reproduce the main experimental results.

\item {\bf Open access to data and code}
    \item[] Question: Does the paper provide open access to the data and code, with sufficient instructions to faithfully reproduce the main experimental results, as described in supplemental material?
    \item[] Answer: \answerNo{} 
    \item[] Justification: We will release the code after acceptance.

\item {\bf Experimental setting/details}
    \item[] Question: Does the paper specify all the training and test details (e.g., data splits, hyperparameters, how they were chosen, type of optimizer) necessary to understand the results?
    \item[] Answer: \answerYes{} 
    \item[] Justification: We have provided the core experimental setup, hardware environment, and training hyperparameters, and explicitly indicated that more comprehensive details are included in the appendix.

\item {\bf Experiment statistical significance}
    \item[] Question: Does the paper report error bars suitably and correctly defined or other appropriate information about the statistical significance of the experiments?
    \item[] Answer: \answerNo{} 
    \item[] Justification: The experimental result for rendering and reconstruction tasks from dense views are very stable, the standard deviation is negligible, so we follow the previous work’s practice and not report the error bar.

\item {\bf Experiments compute resources}
    \item[] Question: For each experiment, does the paper provide sufficient information on the computer resources (type of compute workers, memory, time of execution) needed to reproduce the experiments?
    \item[] Answer: \answerYes{} 
    \item[] Justification: We provide sufficient information on the computer resources, including the type of compute workers GPU.
    
\item {\bf Code of ethics}
    \item[] Question: Does the research conducted in the paper conform, in every respect, with the NeurIPS Code of Ethics \url{https://neurips.cc/public/EthicsGuidelines}?
    \item[] Answer: \answerYes{} 
    \item[] Justification: We have carefully reviewed the NeurIPS Code of Ethics and are committed to strictly adhering to its guidelines.

\item {\bf Broader impacts}
    \item[] Question: Does the paper discuss both potential positive societal impacts and negative societal impacts of the work performed?
    \item[] Answer: \answerNA{} 
    \item[] Justification: Neural rendering and 3D reconstruction are crucial for downstream applications in VR, AR, and game engines. Because our framework successfully outputs lightweight, high-fidelity meshes that directly integrate into traditional graphics pipelines, we believe it has the strong potential to advance progress in this domain and support real-world deployments.
    
\item {\bf Safeguards}
    \item[] Question: Does the paper describe safeguards that have been put in place for responsible release of data or models that have a high risk for misuse (e.g., pre-trained language models, image generators, or scraped datasets)?
    \item[] Answer: \answerNA{} 
    \item[] Justification: There are no such risks in the rendering and reconstruction tasks.

\item {\bf Licenses for existing assets}
    \item[] Question: Are the creators or original owners of assets (e.g., code, data, models), used in the paper, properly credited and are the license and terms of use explicitly mentioned and properly respected?
    \item[] Answer: \answerYes{} 
    \item[] Justification: In the Experiments section, we cite all used code package and public datasets.

\item {\bf New assets}
    \item[] Question: Are new assets introduced in the paper well documented and is the documentation provided alongside the assets?
    \item[] Answer: \answerNo{} 
    \item[] Justification: We will release the code after acceptance.

\item {\bf Crowdsourcing and research with human subjects}
    \item[] Question: For crowdsourcing experiments and research with human subjects, does the paper include the full text of instructions given to participants and screenshots, if applicable, as well as details about compensation (if any)? 
    \item[] Answer: \answerNA{} 
    \item[] Justification: This paper does not involve any crowdsourcing or research with human subjects.

\item {\bf Institutional review board (IRB) approvals or equivalent for research with human subjects}
    \item[] Question: Does the paper describe potential risks incurred by study participants, whether such risks were disclosed to the subjects, and whether Institutional Review Board (IRB) approvals (or an equivalent approval/review based on the requirements of your country or institution) were obtained?
    \item[] Answer: \answerNA{} 
    \item[] Justification: This paper does not involve any crowdsourcing or research with human subjects.

\item {\bf Declaration of LLM usage}
    \item[] Question: Does the paper describe the usage of LLMs if it is an important, original, or non-standard component of the core methods in this research? Note that if the LLM is used only for writing, editing, or formatting purposes and does \emph{not} impact the core methodology, scientific rigor, or originality of the research, declaration is not required.
    \item[] Answer: \answerNA{} 
    \item[] Justification: We only use the LLM for writing, editing, and formatting purposes, which does not affect the core methodology.

\end{enumerate}
\endgroup
\end{document}